\documentclass[runningheads]{llncs}

\usepackage{eccv}

\usepackage{eccvabbrv}
\usepackage{float}
\usepackage{graphicx}
\usepackage{booktabs}

\usepackage[accsupp]{axessibility}  

\usepackage{hyperref}

\makeatletter
\newcommand{\printfnsymbol}[1]{%
  \textsuperscript{\@fnsymbol{#1}}%
}
\makeatother

\begin{document}

\title{AIM 2024 Challenge on Video Saliency Prediction: Methods and Results} 


\author{Andrey Moskalenko \and Alexey Bryncev \and Dmitry Vatolin \and Radu Timofte \and\\ Gen Zhan \and Li Yang \and Yunlong Tang \and Yiting Liao \and Jiongzhi Lin \and Baitao Huang \and Morteza Moradi \and Mohammad Moradi \and Francesco Rundo \and Concetto Spampinato \and Ali Borji \and Simone Palazzo \and Yuxin Zhu \and Yinan Sun \and Huiyu Duan \and Yuqin Cao \and Ziheng Jia \and Qiang Hu \and Xiongkuo Min \and Guangtao Zhai \and Hao Fang \and\\ Runmin Cong \and Xiankai Lu \and Xiaofei Zhou \and Wei Zhang \and Chunyu Zhao \and\\ Wentao Mu \and Tao Deng  \and Hamed R. Tavakoli \thanks{\scriptsize{A.~Moskalenko (and.v.moskalenko@gmail.com), A.~Bryncev (alxbrc0@gmail.com), D.~Vatolin (dmitriy@graphics.cs.msu.ru), and R.~Timofte (radu.timofte@uni-wuerzburg.de) were the challenge organizers, while the other authors participated in the challenge. \cref{affilations} contains the author's teams and affiliations. AIM 2024 webpage: \url{https://www.cvlai.net/aim/2024/}}}}

\authorrunning{A.~Moskalenko, A.~Bryncev, D.~Vatolin, R.~Timofte et al.}

\institute{}

\maketitle

\begin{abstract}

This paper reviews the Challenge on Video Saliency Prediction at AIM 2024. The goal of the participants was to develop a method for predicting accurate saliency maps for the provided set of video sequences. Saliency maps are widely exploited in various applications, including video compression, quality assessment, visual perception studies, the advertising industry, etc. For this competition, a previously unused large-scale audio-visual mouse saliency (\textit{AViMoS}) dataset of 1500 videos with more than 70 observers per video was collected using crowdsourced mouse tracking. The dataset collection methodology has been validated using conventional eye-tracking data and has shown high consistency. Over 30 teams registered in the challenge, and there are 7 teams that submitted the results in the final phase. The final phase solutions were tested and ranked by commonly used quality metrics on a private test subset. The results of this evaluation and the descriptions of the solutions are presented in this report. All data, including the private test subset, is made publicly available on the challenge homepage — \url{https://challenges.videoprocessing.ai/challenges/video-saliency-prediction.html}.

\end{abstract}

\section{Introduction}

Saliency prediction aims to model the human visual system (HVS) by replicating the way humans instinctively focus their attention on certain elements within a visual scene, discerning areas of interest from a complex and dynamic environment.

The ability to obtain high-quality saliency maps plays a crucial role in media content manipulation tasks, e.g. saliency-aware compression~\cite{hadizadeh2013saliency,lyudvichenko2017semiautomatic,patel2021saliency}, media content quality assessment~\cite{gu2016saliency,zhang2017study,yang2019sgdnet}, retargeting~\cite{fang2012saliency,ahmadi2021context,miangoleh2023realistic}, etc., as well as applications in neuroscience and cognitive science.

\subsubsection{Saliency Methods}
Early approaches in saliency prediction often relied on low-level visual features such as color, contrast, and texture. Such natural scene statistics are widely used in many classic methods\cite{itti1998model,harel2006graph,judd2009learning} for predicting saliency maps. Early video saliency prediction methods~\cite{guo2008spatio,mahadevan2009spatiotemporal,marat2009modelling} additionally utilized temporal features (e.g. optical flow) to improve performance and temporal consistency.

Significant progress has been made with the development of deep-learning-based methods in the fields of static~\cite{vig2014large, cornia2018predicting, kroner2020contextual} and dynamic saliency prediction~\cite{bak2017spatio,jiang2018deepvs,min2019tased}. Moreover, some dynamic models also attempt to utilize additional audio modality~\cite{tavakoli2019dave,jain2021vinet,xiong2023casp} for more precise capture of human saliency signal.

\subsubsection{Saliency Data}
Eye-tracking fixations from viewers are commonly used as a source of reference data for saliency prediction. In a laboratory setting, each stimulus is presented to several viewers, while the high-frequency eye-tracking device stores the coordinates of the viewer's fixation points on the screen. Then, the individual fixations are combined into a fixation map and blurred with a Gaussian to obtain a continuous saliency map. The largest video datasets with eye fixations at the moment are Hollywood2~\cite{mathe2014actions} and DHF1K~\cite{wang2018revisiting} with 1707 videos/19 viewers, and  1000 videos/17 viewers respectively.

However, the need for a large number of viewers and a wide variety of possible content creates challenges for collecting large representative datasets. Many existing studies have focused on more scalable ways (e.g. in a crowdsourcing scenario) to collect ground-truth data without an eye-tracker. Mainly, the researchers' interest was attracted by mouse~\cite{rudoy2012crowdsourcing,jiang2015salicon,tavakoli2017saliency,kim2017bubbleview,lyudvichenko2019predicting,Payne2023} and webcam~\cite{xu2015turkergaze,papoutsaki2017searchgazer} tracking. To collect the dataset for this challenge, we based our methodology on the mentioned works that use mouse tracking to gather saliency data. The dataset collection procedure is described in more detail in the \cref{dataset}.

This challenge is one of the AIM 2024 Workshop\footnote{\url{https://www.cvlai.net/aim/2024/}} associated challenges on: sparse neural rendering~\cite{aim2024snr, aim2024snr_dataset}, UHD blind photo quality assessment~\cite{aim2024uhdbpqa}, compressed depth map super-resolution and restoration~\cite{aim2024cdmsrr}, raw burst alignment~\cite{aim2024rawburst}, efficient video super-resolution for AV1 compressed content~\cite{aim2024evsr}, video super-resolution quality assessment~\cite{aim2024vsrqa}, and compressed video quality assessment~\cite{aim2024cvqa}.

\section{AIM 2024 Video Saliency Prediction Challenge}

\subsection{Challenge Data} \label{dataset}

\subsubsection{Data sources}
The challenge data was collected from two sources. As a first data source, we used 246 videos that matched our criteria (e.g. resolution, quality, duration, non-explicit content) from the YouTube-UGC~\cite{wang2019youtube} dataset. We chose the second part of 1254 videos from a pool of more than 15,000 crawled high-bitrate open-source videos from \url{www.vimeo.com}. Our search based on a variety of minor keywords to provide maximum coverage of potential results — for example, “the”, “a”,  “of”, “in”, etc. From both data sources, we downloaded only videos available under CC-BY and CC0 licenses. Additional filtering left only streams with a minimum bitrate of 20 Mbps with at least FullHD resolution. Then all videos were transcoded with libx264 codec with CRF 23 at 30 FPS and downsampled to FullHD resolution (all videos had the same aspect ratio, but could have horizontal/vertical orientation). Additionally, audio streams were normalized according to EBU R128 with transcoding to AAC stereo with 256Kbps. From each video, the assessors from the organizer's team saw a fragment of 15–21 seconds in length (based on the scene change detector timestamps), which was passed to the next phase if it met the quality criteria.

\subsubsection{Saliency collection}

We combined and expanded previous studies and best practices of crowdsourcing saliency data collection to obtain \textit{AViMoS} dataset comparable in quality to an eye-tracker: 

\begin{itemize}

    \item Following~\cite{jiang2015salicon,lyudvichenko2019predicting}, each participant sees a blurred screen except for the area around the cursor, motivating him to move to saliency areas. Following~\cite{lyudvichenko2019predicting} we set up blurring sigma to 2\% of the participant screen width.

    \item We filtered users with low-resolution screens (less than $1280\times720$px). Each video was automatically resized in the browser to fill the maximum screen area without losing the aspect ratio. Further, all collected fixations were resampled to the native video resolution. All views were conducted strictly in full-screen mode to minimize external distractions.

    \item Most of the videos in the challenge dataset contain an audio track, which could potentially influence the distribution of saliency during viewing. Therefore, to make sure that all performers were watching the videos with sound enabled, at the beginning of the experiment, as well as in the middle, the performers had to enter the result of the audio captcha (which was played in their native language).

    \item Before the experiment, all performers took a reaction speed test. The test consists of 3 attempts while a rectangle moves at a constant speed along the perimeter of the screen. The rectangle makes a full rotation inside the screen in 7 seconds, to pass this step the participant should keep the cursor inside it for at least 30\% of the time, which could be problematic for users with low sensitivity or a trackpad.

    \item Each performer watched 23 random videos, of which 3 videos were validation ones (participants did not know which videos were validation, resolution in the validation dataset matches FullHD). Validation was manually selected from the SAVAM~\cite{savam} eye-tracking dataset according to the criteria — there must be one unambiguous source of attention, for some videos it should not coincide with the center of the screen. At this stage, about 5\% of participants with a $CC<0.35$ of their mouse movements w.r.t. ground-truth eye-tracker on any validation video were filtered out.

    \item Following \cite{Payne2023,mital2011clustering} to add interactivity and motivation for attending to the videos, we asked participants how much they liked the video (from 1 to 5 stars) after each view.

    \item Since different browsers and hardware can provide different mouse move event update loop speed, following~\cite{jiang2015salicon} we resampled each mouse movement track with 100Hz frequency by linear interpolation. We also filtered out about 5.5\% of views that had low frequencies (<3Hz).

    \item Using the proposed methodology, the cross-validation dataset~\cite{savam} with eye-tracking data was also fully labeled with mouse tracking. Since mouse movements obviously lag behind eye movements, we found the optimal shift (300ms) through cross-validation with the eye-tracking dataset and applied it to the challenge dataset. 

    \item To further improve data quality, we found optimal trimming time (the first second) and applied it to all videos and annotations as well. The motivation for this step can be explained by the fact that participants need time to initially navigate the mouse to the salient area while viewing. We combined all the cuts, shifts, fps and resolution alignments into one \textit{ffmpeg} command to avoid additional transcoding. Thus, all video streams were transcoded only once from sources.

    \item After all the filtering steps, the obtained metrics of proximity to the eye-tracker data were significantly higher (e.g. $AUC\mathit{-}Judd>0.91$, $CC>0.84$, $SIM>0.74$) than the \href{https://videoprocessing.ai/benchmarks/video-saliency-prediction.html}{\color{eccvblue}{results}} of the state-of-the-art automatic methods on the same dataset~\cite{savam}. This empirically justifies that the obtained data can be used as a ground-truth for conducting the saliency prediction challenge.

    \item All stages of filtering were successfully passed by \textbf{>5000 participants}, providing on average \textbf{>70 unique viewers} for each of the \textbf{1500 videos} with mean \textbf{19s} duration.

\end{itemize}

\noindent To generate continuous saliency maps from the obtained fixation maps, we gathered all the fixations corresponding to each frame timestamp interval and applied Gaussian with the $1920\cdot0.02 = 38.4px$ sigma to match it with the mouse viewer's blur sigma during the data collection phase~\cite{lyudvichenko2019predicting}.

\begin{table}
\fontsize{10pt}{13.35pt}\selectfont
\tabcolsep=4.25pt
\begin{tabular}{clccccccc}
\toprule
\multicolumn{1}{l}{} &  & \multicolumn{4}{c}{Private Test Subset Metrics}                                                                                          &                      & \multicolumn{2}{c}{Additional info} \\
Team Name            &  & AUC-Judd                       & CC                             & SIM                            & NSS                            &                      & Rank       & \#Params(M)      \\ \cline{1-1} \cline{3-6} \cline{8-9} 
CV\_MM               &  & \textbf{0.894}                 & \textbf{0.774}                 & \textbf{0.635}                 & \textbf{3.464}                 &                      & 1.00            & 420.5             \\
VistaHL              &  & \underline{0.892} & \underline{0.769} & \underline{0.623} & 3.352                          &                      & 2.75            & 187.7             \\
PeRCeiVe Lab         &  & 0.857                          & \textit{0.766}                 & 0.610                          & \underline{3.422} &                      & 3.75            & 402.9             \\
SJTU-MML             &  & 0.858                          & 0.760                          & \textit{0.615}                 & 3.356                          &                      & 4.00            & 1288.7            \\
MVP                  &  & 0.838                          & 0.749                          & 0.587                          & \textit{3.404}                 &                      & 5.00            & 99.6              \\
ZenithChaser         &  & \textit{0.869}                 & 0.606                          & 0.517                          & 2.482                          &                      & 5.50            & 0.19              \\
Exodus               &  & 0.861                          & 0.599                          & 0.510                          & 2.491                          & \multicolumn{1}{l}{} & 6.00            & 69.7              \\
Baseline             &  & 0.833                          & 0.449                          & 0.424                          & 1.659                          &                      & 8.00            & —  \\
\bottomrule
\end{tabular}
\caption{Results of AIM 2024 Video Saliency Prediction Challenge. Best scores are shown in \textbf{bold}, the second best is \underline{underlined}, while the third best is \textit{italic}. The ranking is based on the mean rank across all the metrics (\cref{evaluation}). The \#Params column describes for each model the number of parameters in millions.}
\label{table:results}
\end{table}

\subsection{Evaluation} \label{evaluation}

To objectively assess the similarity between methods predictions and ground-truth, we used 4 common metrics~\cite{riche2013saliency,bylinskii2018different} — Area Under the Curve (AUC-Judd), Pearson’s Correlation Coefficient (CC), Similarity or histogram intersection (SIM), and Normalized Scanpath Saliency (NSS). The final rank for a participant is calculated as the average rank on all four metrics on the test set. If the final rank is equal by different methods, the result of the first non-matching metric in the order they are listed. In the final phase, participants provided the final predictions, factsheet, and code to reproduce the submitted results, which were validated by the organizers. In total, 7 teams successfully passed this phase. Additionally, the final evaluation contains the organizers' team baseline solution, which was available to all participants as a sample submission.

\subsection{Challenge phases}
The dataset was randomly split into 2 parts in a 2:1 ratio — training (1000 videos with fixations and saliency maps) and testing (500 videos) subsets. The test subset was randomly divided into two parts — a validation part for online public testing (150 videos) and a private test part (350 videos). During the competition, participants were able to make submissions and see their results on a first subset of 150 videos. In the final phase, methods were tested on a hidden private test subset. Ground-truth saliency maps and fixations for the test subset became available only after the end of the competition.

\begin{table}
\fontsize{10pt}{14pt}\selectfont
\tabcolsep=3.95pt
\begin{tabular}{clccccccc}
\toprule
\multicolumn{1}{l}{} &  & \multicolumn{4}{c}{Public Test Subset Metrics}                                                                &                      & \multicolumn{2}{c}{Additional info} \\
Team Name            &  & AUC-Judd       & CC                                                   & SIM            & NSS            &                      & Rank       & \#Params(M)      \\ \cline{1-1} \cline{3-6} \cline{8-9} 
CV\_MM               &  & \textbf{0.894} & \underline{0.7738}                                         & \textbf{0.633} & \textbf{3.485} &                      & 1.25            & 420.5             \\
VistaHL              &  & \underline{0.892}    & \textbf{0.7740}                                      & \underline{0.625}    & 3.411          &                      & 2.25            & 187.7             \\
PeRCeiVe Lab         &  & 0.853          & \textit{0.768}                                       & 0.608          & \textit{3.464} &                      & 4.00            & 402.9             \\
SJTU-MML             &  & 0.854          & 0.761                                                & \textit{0.614} & 3.396          &                      & 4.25            & 1288.7            \\
MVP                  &  & 0.834          & 0.757                                                & 0.589          & \underline{3.477}    &                      & 5.00            & 99.6              \\
ZenithChaser         &  & \textit{0.871} & 0.623                                                & 0.527          & 2.596          &                      & 5.25            & 0.19              \\
Exodus               &  & 0.859          & 0.599 & 0.509          & 2.505          & \multicolumn{1}{l}{} & 6.25            & 69.7              \\
Baseline             &  & 0.837          & 0.455                                                & 0.424          & 1.688          &                      & 7.75            & —  \\
\bottomrule
\end{tabular}
\caption{Public Test subset results. Participants did not see the ground-truth of this subset until the end of the challenge, however, they could make online submissions (up to 5 times a day) to get scores on this subset and observe the preliminary leaderboard. Best scores are shown in \textbf{bold}, the second best is \underline{underlined}, while the third best is \textit{italic}. The ranking is based on the mean rank across all the metrics (\cref{evaluation}). The \#Params column describes for each model the number of parameters in millions.}
\label{table:public_results}
\end{table}

\section{Results}

This section introduces the results of AIM 2024 Video Saliency Prediction Challenge. The values of all metrics and the final ranking are presented in ~\cref{table:results}. The public test set results are presented in \cref{table:public_results}. Top solutions utilized Transformer-based encoders to extract video features. The first-place team in the rankings (CV\_MM,~\cref{cv_mm_solution}) has employed the UMT model~\cite{li2023unmasked} and adopted features of different resolutions in the decoder phase. The team in the second place (VistaHL,~\cref{vistahl_solution}) created an architecture with two branches by encoding the low-resolution video context as well as the high-resolution context of the current frame. The third place team (PeRCeiVe Lab,~\cref{PeRCeiVe_solution}) applied UMT~\cite{li2023unmasked} with multiple decoding branches to focus on different spatio-temporal saliency information sources. Next, the SJTU-MML team (\cref{SJTU_MML_solution}) additionally used audio information and applied transformer blocks both to the input frames and to the sound Mel-spectrograms. The MVP team (\cref{MVP_solution}) adopted Video Swin Transformer~\cite{liu2022video} to extract spatio-temporal features from different resolutions and aggregate them using 3D convolutions. The ZenithChaser team (\cref{ZenithChaser_solution}) proposed an extremely light-weight solution based on Mamba~\cite{gu2023mamba,dao2024transformers}, archiving efficient saliency prediction model. The Exodus team (\cref{Exodus_solution}) used a two-branch model~\cite{tavakoli2019dave} based on 3D convolutions, where one branch processed frames while the other processed the Mel-spectrograms of the audio.

\section{Teams solutions}

\subsection{CV\_MM}\label{cv_mm_solution}
\begin{figure}
    \centering
    \includegraphics[width=1.0\textwidth]{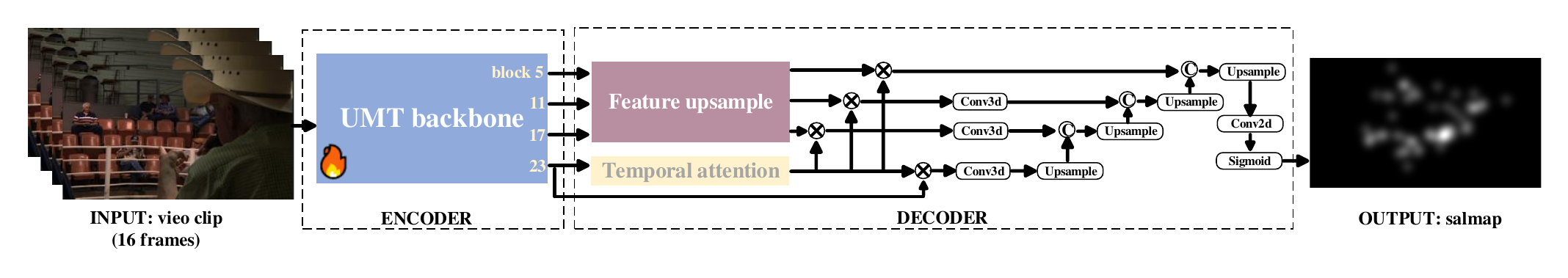}
    \caption{RPN for video saliency prediction.}
    \label{fig:cv_mm}
\end{figure}

In this contest, we propose a new video saliency prediction (VSP) model on the basis of an encoder-decoder deep learning network. The encoder is used for extracting underlying spatio-temporal features in light of the pretrained UMT model (Unmasked teacher: Towards training-efficient video foundation models \cite{li2023unmasked}). With the multi-level features from the encoder, the decoder integrates each feature hierarchically in a top-down manner and finally generates the saliency maps. The pipeline of the proposed method is provided in \cref{fig:cv_mm}.

\subsection{VistaHL}\label{vistahl_solution}

Recent video saliency prediction (VSP) methods based on deep neural networks have achieved remarkable performance. However, due to the limitations imposed by the computational complexity of the models, most existing models use low-resolution videos as input to predict video saliency. To overcome this limitation, we experimentally introduce high-resolution video frame as additional input and propose a new dual-stream framework (HiSal). This framework consists of a spatio-temporal semantic encoding branch based on dense low-resolution video frames and a spatial detail encoding branch based on a single high-resolution video frame. The low-resolution branch extracts spatio-temporal features using a Transformer Backbone and then transfers these features to guide the encoding of the high-resolution branch. To effectively utilize the features from the low-resolution branch, we propose a Selective Cross Attention Module (SCAM), which enables the high-resolution branch to select corresponding saliency regions for feature extraction. 

\begin{figure}[tb]
    \centering
    \includegraphics[width=1.0\textwidth]{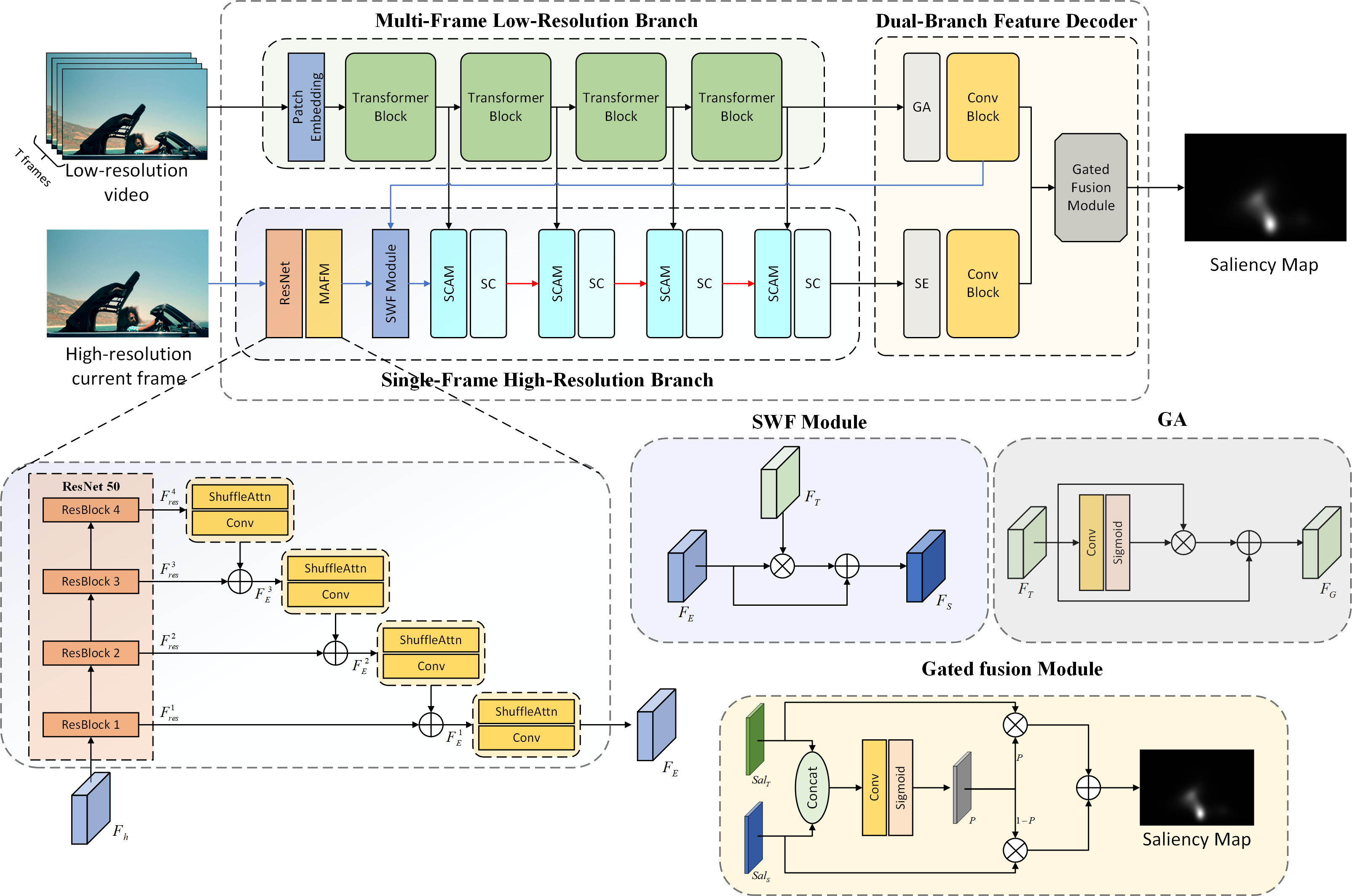}
    \caption{An overview of the proposed network. SC\cite{li2023scconv}, SE\cite{hu2018squeeze}, and ShuffleAttn\cite{zhang2021sa} are plug-and-play attention modules. SWF and GA stand for Saliency-Weighted Feature Module and Gated Attention, respectively.}
    \label{fig:vistahl}
\end{figure}

This avoids the problem of inefficient computation caused by the imbalance between saliency and non-saliency regions in high-resolution video frame, as well as issues with noise interference. 
Additionally, introducing high-resolution frames for saliency prediction inevitably brings a large amount of redundant information. We address this by employing a plug-and-play attention module as a filter to eliminate redundancy. Furthermore, we design a Saliency-Weighted Feature Module (SWF), which uses the saliency mask generated from the low-resolution branch to explicitly enhance saliency feature in the high-resolution branch.

The overall method is illustrated in the Fig.~\ref{fig:vistahl}. The network consists of three parts: the Multi-Frame Low-Resolution Branch, the Single-Frame High-Resolution Branch, and the Dual-Branch Feature Decoder. 

\begin{figure}[tb]
    \centering
    \includegraphics[width=1.0\textwidth]{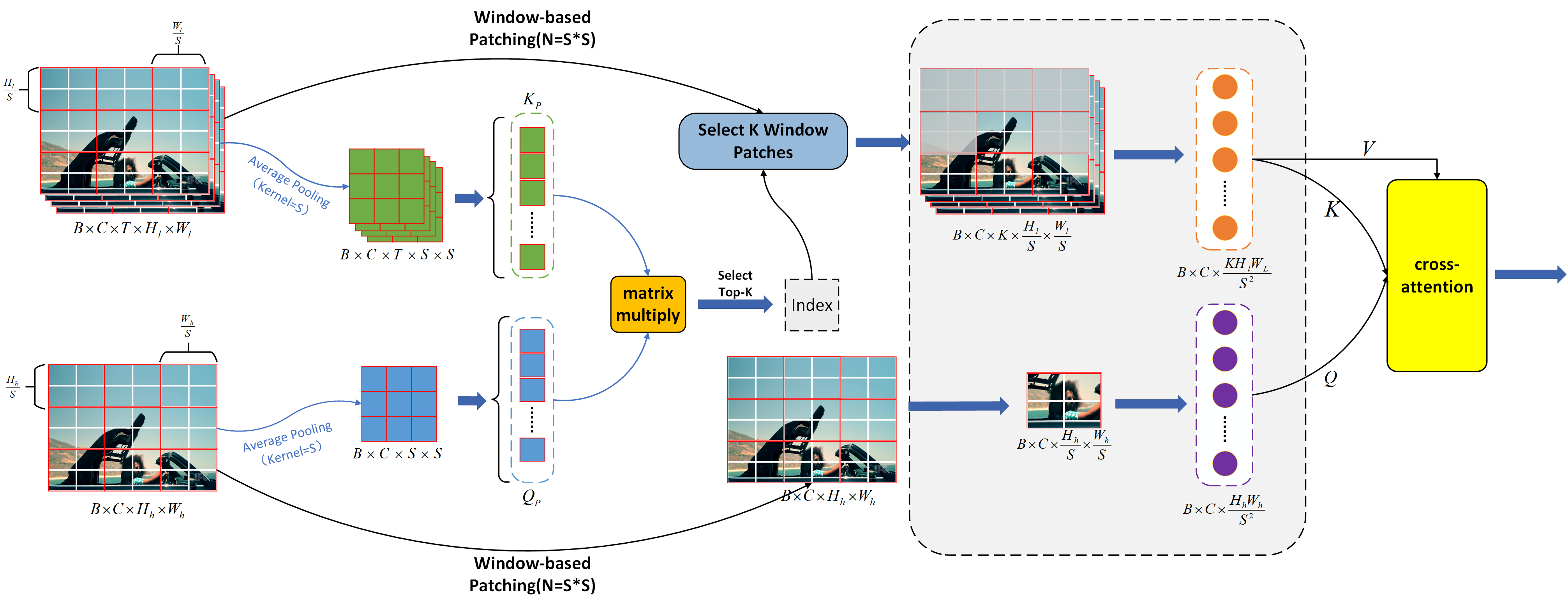}
    \caption{An overview of the Selective Cross Attention Module (SCAM).}
    \label{fig:SCAM}
\end{figure}

The Selective Cross Attention Module (SCAM) is shown in the Fig.~\ref{fig:SCAM}. Features from the two branches are pooled to compute attention scores. The indices of top K attention scores are selected to guide the Selective Cross Attention calculation of the features from the two branches.

\subsection{PeRCeiVe Lab}\label{PeRCeiVe_solution}
\begin{figure}[tb]
    \centering
    \includegraphics[width=1.0\textwidth]{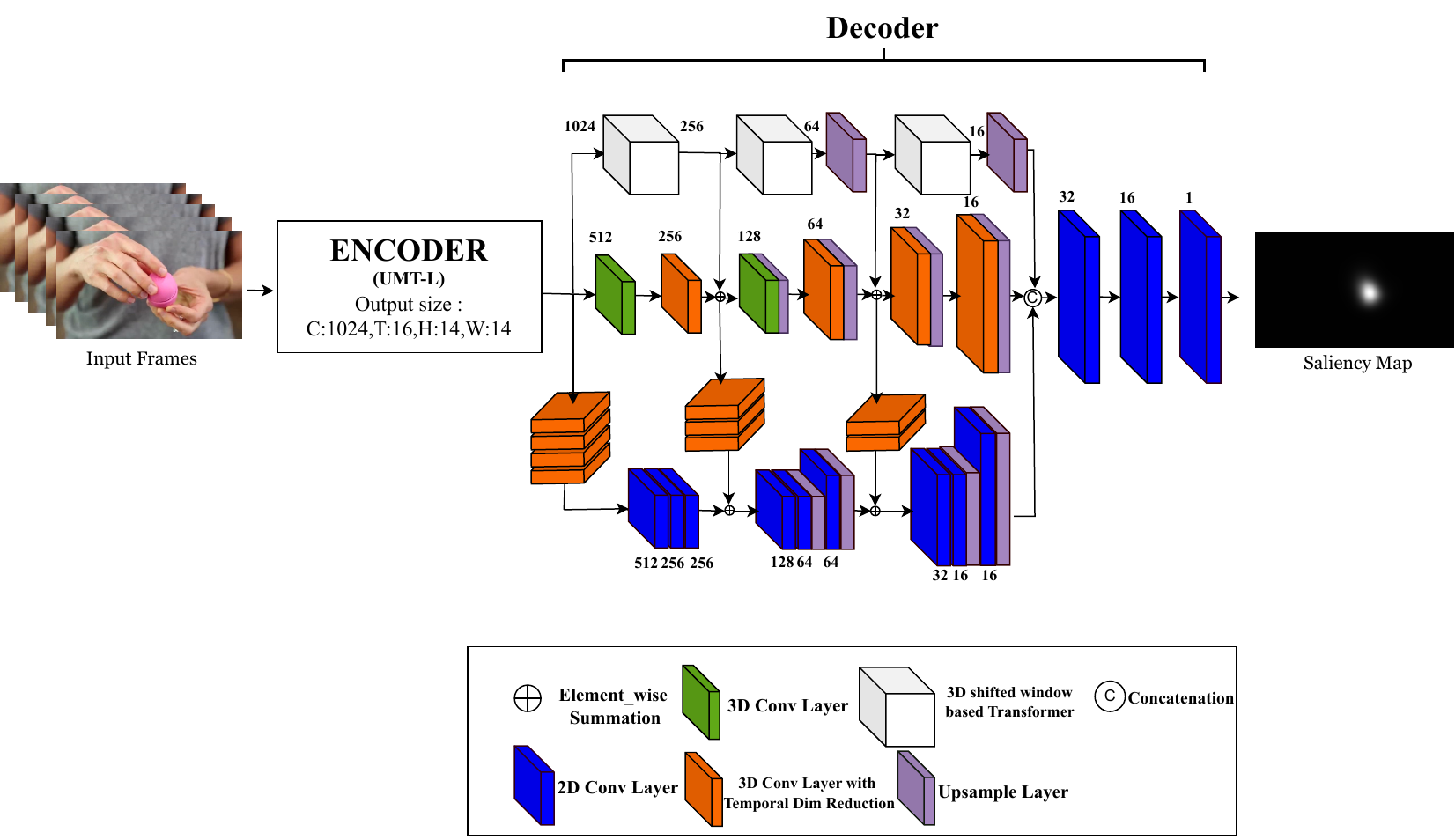}
    \caption{Summary of the architecture of SalFoM. }
    \label{fig:PeRCeiVe_network}
\end{figure}

SalFoM \cite{salfom} is a video saliency prediction model using a video foundation model (VFM) as its feature encoder and a heterogeneous decoder. The multiperspective heterogeneous decoder captures and integrates diverse aspects of spatio-temporal information from the VFM encoder to ensure a comprehensive saliency map in both space and time, crucial for attention modeling in videos. Inspired by strategies that maintain temporal resolution close to the original input, SalFoM gradually reduces the temporal dimension to avoid abrupt loss of information. It leverages the VFM encoder's expressive features, aiming for effective feature analysis and interaction rather than extracting complex features. This involves reducing the channel dimension to a compact representation, facilitating efficient computation and potentially improving generalization. The decoder network includes the Transformer-based Complementary Feature Extraction (TCFE) branch, which captures spatio-temporal relationships and encodes them into feature maps. The Dynamic Feature Decoding (DFD) branch focuses on maintaining temporally-rich information and extracting detailed local features, gradually increasing and recovering the original input resolution. The Static Feature Decoding (SFD) branch abstracts temporal effects to focus on spatial information, recognizing that not all temporal information is relevant for saliency prediction. The final feature fusion stage integrates features from all intermediate branches, producing the final saliency map through 2D convolutional layers. The summary of the architecture is provided in \cref{fig:PeRCeiVe_network}.

\begin{figure}[h!]
    \centering
    \includegraphics[width=1.0\textwidth]{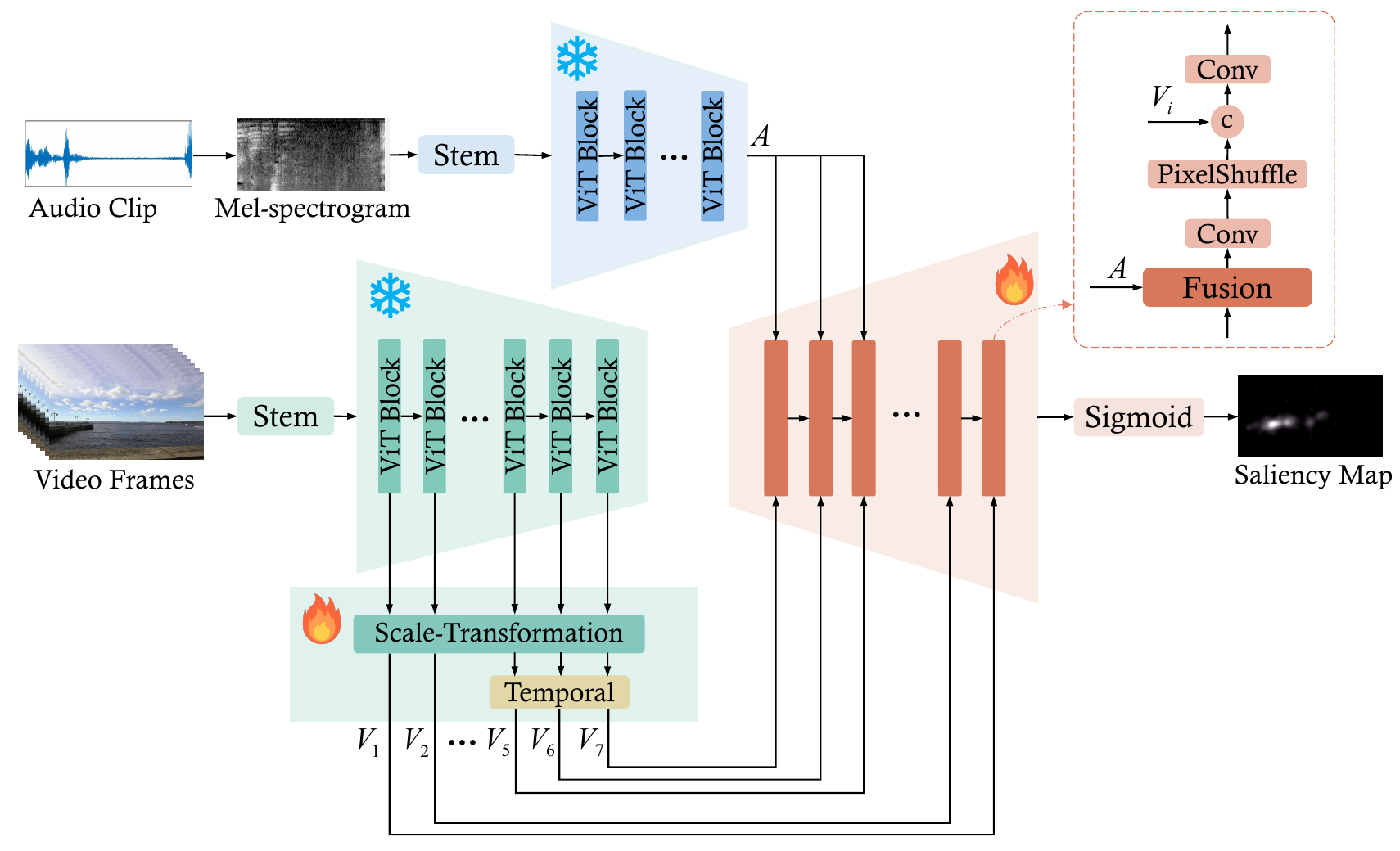}
    \caption{A detailed illustration of our proposed AVSal framework.}
    \label{fig:model_sjtu}
\end{figure}

\subsection{SJTU-MML}\label{SJTU_MML_solution}
As illustrated in Fig. \ref{fig:model_sjtu}, we developed an Audio-Visual Saliency prediction network (AVSal) based on the U-Net architecture. The AVSal framework consists of a spatial visual feature representation module that extracts spatial characteristics from video frames, a visual feature temporal aggregation module that integrates temporal information, an audio feature representation module for extracting semantic features from audio clips, an audio-visual feature fusion module that hierarchically infuses audio features into visual features, and audio-visual saliency estimation blocks that decode multi-scale audio-visual features to generate saliency maps.

\begin{figure}[b!]
    \centering
    \includegraphics[width=1.0\textwidth]{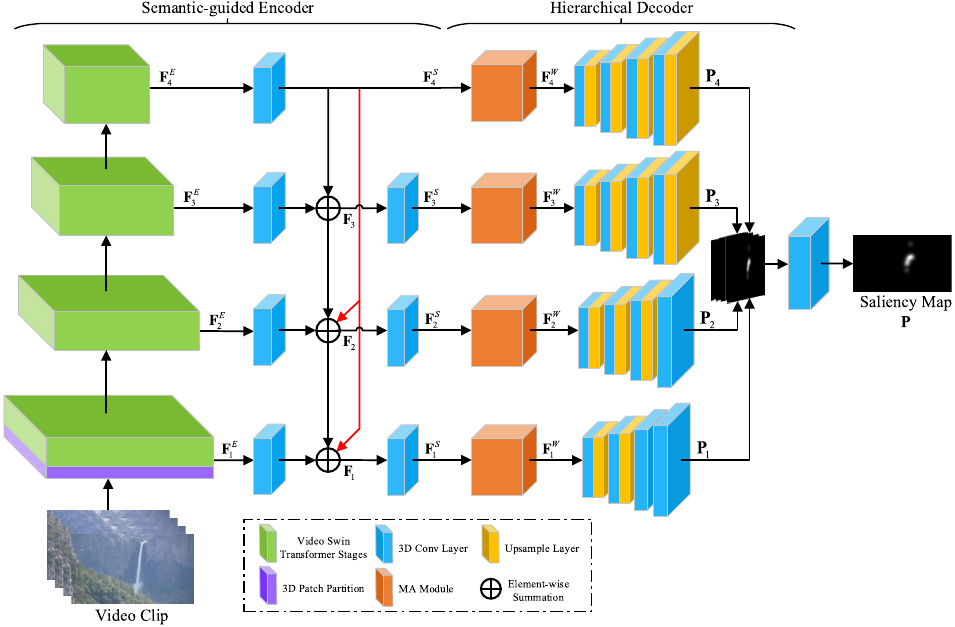}
    \caption{The overall architecture of TMFI-Net.}
    \label{fig:mvp}
\end{figure}

\subsection{MVP}\label{MVP_solution}
As shown in \cref{fig:mvp}, our pipeline is based on TMFI-Net\cite{zhou2023transformer}, and we make modifications to both the backbone and decoder. TMFI-Net is built on the Video Swin Transformer\cite{liu2022video}. Specifically, firstly, video clips are sent to TMFI-Net, and we can obtain multi-level spatiotemporal features $\left \{ F_{i}^{E}  \right \} _{i=1}^{4}$. Then, the semantic-guided encoder continuously integrates high-level features with low-level features via a top-down pathway and longitudinal connection, which obtain multi-scale semantic features $\left \{ F_{i}^{S}  \right \} _{i=1}^{4}$. After that, the hierarchical decoder deploys multi-dimensional attention (MA) module to accurately locate salient regions and remove redundant information, generating multi-level weighted features $\left \{ F_{i}^{W}  \right \} _{i=1}^{4}$. Subsequently, four decoding blocks generate four coarse saliency maps  $\left \{ P_{i} \right \} _{i=1}^{4}$. Finally, the four saliency maps are combined, where a 3D convolutional layer is used to generate the final saliency map $P$. Unlike TMFI-Net, we use Swin-B instead of Swin-S weight for training. To preserve more multi-scale information, we directly concatenate multi-level weighted features$\left \{ F_{i}^{W}  \right \} _{i=1}^{4}$ and then use a decoding block to output the final saliency map. Our model is trained on the training set of DHF1K~\cite{wang2018revisiting} and fine-tune on the training set of challenge dataset.

\subsection{ZenithChaser}\label{ZenithChaser_solution}
\begin{figure}[tb]
    \centering
    \includegraphics[width=1.0\textwidth]{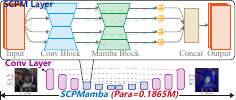}
    \caption{Model Illustration: SCPMamba pipeline scheme.}
    \label{fig:pipeline}
\end{figure}

In our scheme, we construct an extremely lightweight pipeline containing only $\textbf{0.1865M}$ parameters and with a model size of only $\textbf{2.901MB}$, the pipeline is shown in Fig.~\ref{fig:pipeline}. In this pipeline, we first resize the input image to $512^2$, which is first fed into the convolutional layer, and then further extract features using a selective channel parallel mamba ($SCPM$) layer. The $SCPM$ layer which embeds the input features in the convolutional block in parallel and then feeds them into the mamba layer in parallel. The $SCPM$ layer not only solves the parametric catastrophe brought by high-dimensional data to mamba, but also corrects the information loss brought by fixed channel slicing.

\subsection{Exodus}\label{Exodus_solution}

\begin{figure}[tb]
	\centering
	\includegraphics[width=1.0\textwidth]{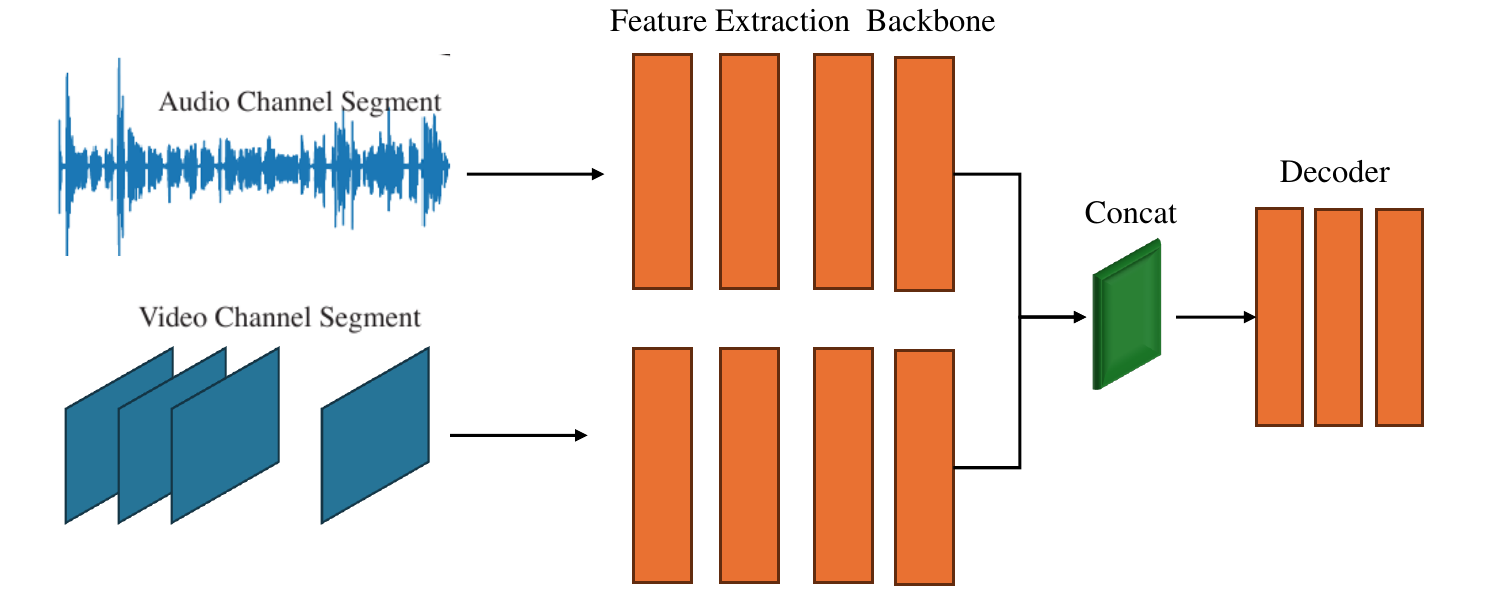}
	\caption{Simplified DAVE architecture. Audio and Video information are fed into a feature extraction backbone, which is based on 3D ResNet blocks. The extracted features are concatenated and fed into a decoder. The decoder consists of 1x1 convolutions and up-sampling blocks. 
	}
	\label{fig:exodus}
\end{figure}

We investigate a multimodal saliency model, aka DAVE~\cite{tavakoli2019dave}. The saliency prediction architecture is depicted in Figure~\ref{fig:exodus}. It consists of a two-stream neural network-based feature extraction, each stream corresponding to one modality. The feature extractor follows the ResNet18 3D Convolutional neural networks (3D CNNs) architecture.  The video branch input is of size $F\times C\times256\times320$, where $F=16$ is the number of frames and $C=3$ is the number of channels. The audio signal is provided as log mel-spectrogram with a window length of 0.025 seconds and a hop length of 0.01 seconds with 64 bands. We then convert the transformed audio information into a sequence of successive overlapping frames, resulting in an audio tensor representation of shape $F\times C\times 64 \times 64$, where $C=1$ is the number of channels. The extracted audio and video features are concatenated together. The extracted features are fed into a saliency decoder which consists of 1x1 2D convolution layers and bilinear up-sampling layers with factor of 2. The decoder produces saliency maps of size  $32 \times 40$. The model is trained using KL-divergence as loss function. For the final submission, the solution utilizes the model weights that are available at \href{https://github.com/hrtavakoli/DAVE}{\color{eccvblue}{GitHub}}.

\subsection{Center Prior Baseline}
This solution was provided by the organizers team as a baseline. To obtain it, the average saliency map was calculated for all frames of all videos from the training set (e.g. 1000 videos). The resulting single-channel map was fitted with a Gaussian centered at the geometric center of the frame, while only the $\sigma_x$ and $\sigma_y$ distribution parameters were optimized to minimize L2-Norm. The resulting center prior frame was replicated for each frame for each test video. This solution was available for download to all participants as a sample submission.

\section{Conclusion}
In the AIM 2024 Video Saliency Prediction challenge, 7 teams competed to develop state-of-the-art video saliency prediction methods with the previously undisclosed \textit{AViMoS} dataset. Most final solutions used Transformer-based architectures and tried to utilize spatio-temporal information as much as possible. Moreover, two teams additionally adopted information from the audio stream as well. In conclusion, we would like to note that despite the existence of the saliency prediction task for many decades, this task remains unsolved and competitive.

\section{Teams and Affiliations}\label{affilations}
\subsection{CV\_MM}
\textit{\textbf{Members:}}\\
Gen Zhan$^1$, Li Yang$^1$, Yunlong Tang$^{2,3}$, Yiting Liao$^{1,2}$\\
\textit{\textbf{Affiliations:}}\\
$^1$ ByteDance, China\\
$^2$ ByteDance, USA \\
$^3$ University of Rochester, USA

\subsection{VistaHL}
\textit{\textbf{Members:}}\\
Jiongzhi Lin$^{1,2,3}$ (2019281030@email.szu.edu.cn), \\Baitao Huang$^{1,2,3}$ (2210433076@email.szu.edu.cn)\\
\textit{\textbf{Affiliations:}}\\
$^1$ College of Electronic and Information Engineering, Shenzhen University, China\\
$^2$ Shenzhen Key Laboratory of Digital Creative Technology, China\\
$^3$ Guangdong Key Laboratory of Intelligent Information Processing, China

\subsection{PeRCeiVe Lab}
\textit{\textbf{Members:}}\\
Morteza Moradi$^1$ (morteza.moradi@phd.unict.it),\\ Mohammad Moradi$^1$ (Mohammad.moradi@phd.unict.it),\\ Francesco Rundo$^2$ (francesco.rundo@st.com),\\ Concetto Spampinato$^1$ (concetto.spampinato@unict.it),\\ Ali Borji$^3$ (aliborji@gmail.com),\\ Simone Palazzo$^1$ (simone.palazzo@unict.it) \\
\textit{\textbf{Affiliations:}}\\
$^1$ University of Catania, Italy\\
$^2$ STMicrolectronics, ADG Central R\&D, Italy \\
$^3$ Quintic AI, San Francisco, USA

\subsection{SJTU-MML}
\textit{\textbf{Members:}}\\
Yuxin Zhu$^1$ (rye2000@sjtu.edu.cn), Yinan Sun$^1$  (yinansun@sjtu.edu.cn),\\ Huiyu Duan$^1$ (huiyuduan@sjtu.edu.cn), Yuqin Cao$^1$ (caoyuqin@sjtu.edu.cn),\\ Ziheng Jia$^1$ (jzhwsl@sjtu.edu.cn), Qiang Hu$^1$ (qiang.hu@sjtu.edu.cn),\\ Xiongkuo Min$^1$ (minxiongkuo@sjtu.edu.cn),\\ Guangtao Zhai$^1$ (zhaiguangtao@sjtu.edu.cn)\\
\textit{\textbf{Affiliations:}}\\
$^1$ Shanghai Jiao Tong University, China

\subsection{MVP}
\textit{\textbf{Members:}}\\
Hao Fang$^1$ (fanghaook@mail.sdu.edu.cn), Runmin Cong$^1$ (rmcong@sdu.edu.cn), Xiankai Lu$^{2}$ (carrierlxk@gmail.com), Xiaofei Zhou$^{3}$ (zxforchid@outlook.com), Wei Zhang$^1$ (davidzhang@sdu.edu.cn)\\
\textit{\textbf{Affiliations:}}\\
$^1$ School of Control Science and Engineering, Shandong University, China\\
$^2$ School of Software, Shandong University, China \\
$^3$ School of Automation, Hangzhou Dianzi University, China

\subsection{ZenithChaser}
\textit{\textbf{Members:}}\\
Chunyu Zhao$^1$ (zhaochunyu@my.swjtu.edu.cn),\\ Wentao Mu$^1$ (mwt2858781490@my.swjtu.edu.cn),\\ Tao Deng$^{1}$ (tdeng@swjtu.edu.cn)\\
\textit{\textbf{Affiliations:}}\\
$^1$ School of Information Science and Technology, Southwest Jiaotong University, China

\subsection{Exodus}
\textit{\textbf{Members:}}\\
Hamed R. Tavakoli$^1$ (hamed.rezazadegan\_tavakoli@nokia.com)\\
\textit{\textbf{Affiliations:}}\\
$^1$ Nokia Technologies, Finland

\subsection{Organizers of AIM 2024 Video Saliency Prediction Challenge}
\textit{\textbf{Members:}}\\
Andrey Moskalenko$^{1,2,\dagger}$ (and.v.moskalenko@gmail.com), \\Alexey Bryncev$^{1,\dagger}$ (alxbrc0@gmail.com),\\ Dmitry Vatolin$^{1,3}$ (dmitriy@graphics.cs.msu.ru), \\Radu Timofte$^{4}$ (radu.timofte@uni-wuerzburg.de)\\
\textit{\textbf{Affiliations:}}\\
$^1$ Lomonosov Moscow State University, Russia\\
$^2$ AIRI, Moscow, Russia\\
$^3$ MSU Institute for Artificial Intelligence, Russia\\
$^4$ University of Würzburg, Germany\\
$^\dagger$ Equal contribution

\section*{Acknowledgements}
This work was partially supported by the Humboldt Foundation. We thank the AIM 2024 sponsors: Meta Reality Labs, KuaiShou, Huawei, Sony Interactive Entertainment and University of W\"urzburg (Computer Vision Lab). 

\bibliographystyle{splncs04}
\bibliography{main}

\begin{thebibliography}{10}
\providecommand{\url}[1]{\texttt{#1}}
\providecommand{\urlprefix}{URL }
\providecommand{\doi}[1]{https://doi.org/#1}

\bibitem{ahmadi2021context}
Ahmadi, M., Karimi, N., Samavi, S.: Context-aware saliency detection for image retargeting using convolutional neural networks. Multimedia Tools and Applications  \textbf{80}(8),  11917--11941 (2021)

\bibitem{bak2017spatio}
Bak, C., Kocak, A., Erdem, E., Erdem, A.: Spatio-temporal saliency networks for dynamic saliency prediction. IEEE Transactions on Multimedia  \textbf{20}(7),  1688--1698 (2017)

\bibitem{bylinskii2018different}
Bylinskii, Z., Judd, T., Oliva, A., Torralba, A., Durand, F.: What do different evaluation metrics tell us about saliency models? IEEE transactions on pattern analysis and machine intelligence  \textbf{41}(3),  740--757 (2018)

\bibitem{aim2024rawburst}
Conde, M.V., Bishop, T., Timote, R., Kolmet, M., MacEwan, D., Vinod, V., Tan, J., et~al.: {AIM} 2024 challenge on raw burst alignment via optical flow estimation. In: Proceedings of the European Conference on Computer Vision (ECCV) Workshops (2024)

\bibitem{aim2024evsr}
Conde, M.V., Lei, Z., Li, W., Katsavounidis, I., Timofte, R., et~al.: {AIM} 2024 challenge on efficient video super-resolution for av1 compressed content. In: Proceedings of the European Conference on Computer Vision (ECCV) Workshops (2024)

\bibitem{aim2024cdmsrr}
Conde, M.V., Vasluianu, F.A., Xiong, J., Ye, W., Ranjan, R., Timofte, R., et~al.: Compressed depth map super-resolution and restoration: {AIM} 2024 challenge results. In: Proceedings of the European Conference on Computer Vision (ECCV) Workshops (2024)

\bibitem{cornia2018predicting}
Cornia, M., Baraldi, L., Serra, G., Cucchiara, R.: Predicting human eye fixations via an lstm-based saliency attentive model. IEEE Transactions on Image Processing  \textbf{27}(10),  5142--5154 (2018)

\bibitem{dao2024transformers}
Dao, T., Gu, A.: Transformers are ssms: Generalized models and efficient algorithms through structured state space duality. arXiv preprint arXiv:2405.21060  (2024)

\bibitem{fang2012saliency}
Fang, Y., Chen, Z., Lin, W., Lin, C.W.: Saliency detection in the compressed domain for adaptive image retargeting. IEEE Transactions on Image Processing  \textbf{21}(9),  3888--3901 (2012)

\bibitem{savam}
Gitman, Y., Erofeev, M., Vatolin, D., Bolshakov, A., Fedorov, A.: Semiautomatic {Visual-Attention} modeling and its application to video compression. In: 2014 IEEE International Conference on Image Processing (ICIP) (ICIP 2014). pp. 1105--1109. Paris, France (Oct 2014)

\bibitem{gu2023mamba}
Gu, A., Dao, T.: Mamba: Linear-time sequence modeling with selective state spaces. arXiv preprint arXiv:2312.00752  (2023)

\bibitem{gu2016saliency}
Gu, K., Wang, S., Yang, H., Lin, W., Zhai, G., Yang, X., Zhang, W.: Saliency-guided quality assessment of screen content images. IEEE Transactions on Multimedia  \textbf{18}(6),  1098--1110 (2016)

\bibitem{guo2008spatio}
Guo, C., Ma, Q., Zhang, L.: Spatio-temporal saliency detection using phase spectrum of quaternion fourier transform. In: 2008 IEEE conference on computer vision and pattern recognition. pp.~1--8. IEEE (2008)

\bibitem{hadizadeh2013saliency}
Hadizadeh, H., Baji{\'c}, I.V.: Saliency-aware video compression. IEEE Transactions on Image Processing  \textbf{23}(1),  19--33 (2013)

\bibitem{harel2006graph}
Harel, J., Koch, C., Perona, P.: Graph-based visual saliency. Advances in neural information processing systems  \textbf{19} (2006)

\bibitem{aim2024uhdbpqa}
Hosu, V., Conde, M.V., Timofte, R., Agnolucci, L., Zadtootaghaj, S., Barman, N., et~al.: {AIM} 2024 challenge on uhd blind photo quality assessment. In: Proceedings of the European Conference on Computer Vision (ECCV) Workshops (2024)

\bibitem{hu2018squeeze}
Hu, J., Shen, L., Sun, G.: Squeeze-and-excitation networks. In: Proceedings of the IEEE/CVF Conference on Computer Vision and Pattern Recognition. pp. 7132--7141 (2018)

\bibitem{itti1998model}
Itti, L., Koch, C., Niebur, E.: A model of saliency-based visual attention for rapid scene analysis. IEEE Transactions on pattern analysis and machine intelligence  \textbf{20}(11),  1254--1259 (1998)

\bibitem{jain2021vinet}
Jain, S., Yarlagadda, P., Jyoti, S., Karthik, S., Subramanian, R., Gandhi, V.: Vinet: Pushing the limits of visual modality for audio-visual saliency prediction. In: 2021 IEEE/RSJ International Conference on Intelligent Robots and Systems (IROS). pp. 3520--3527. IEEE (2021)

\bibitem{jiang2018deepvs}
Jiang, L., Xu, M., Liu, T., Qiao, M., Wang, Z.: Deepvs: A deep learning based video saliency prediction approach. In: Proceedings of the European Conference on Computer Vision (ECCV). pp. 602--617 (2018)

\bibitem{jiang2015salicon}
Jiang, M., Huang, S., Duan, J., Zhao, Q.: Salicon: Saliency in context. In: Proceedings of the IEEE conference on computer vision and pattern recognition. pp. 1072--1080 (2015)

\bibitem{judd2009learning}
Judd, T., Ehinger, K., Durand, F., Torralba, A.: Learning to predict where humans look. In: 2009 IEEE 12th international conference on computer vision. pp. 2106--2113. IEEE (2009)

\bibitem{kim2017bubbleview}
Kim, N.W., Bylinskii, Z., Borkin, M.A., Gajos, K.Z., Oliva, A., Durand, F., Pfister, H.: Bubbleview: an interface for crowdsourcing image importance maps and tracking visual attention. ACM Transactions on Computer-Human Interaction (TOCHI)  \textbf{24}(5),  1--40 (2017)

\bibitem{kroner2020contextual}
Kroner, A., Senden, M., Driessens, K., Goebel, R.: Contextual encoder--decoder network for visual saliency prediction. Neural Networks  \textbf{129},  261--270 (2020)

\bibitem{li2023scconv}
Li, J., Wen, Y., He, L.: Scconv: Spatial and channel reconstruction convolution for feature redundancy. In: Proceedings of the IEEE/CVF Conference on Computer Vision and Pattern Recognition. pp. 6153--6162 (2023)

\bibitem{li2023unmasked}
Li, K., Wang, Y., Li, Y., Wang, Y., He, Y., Wang, L., Qiao, Y.: Unmasked teacher: Towards training-efficient video foundation models. In: Proceedings of the IEEE/CVF International Conference on Computer Vision. pp. 19948--19960 (2023)

\bibitem{liu2022video}
Liu, Z., Ning, J., Cao, Y., Wei, Y., Zhang, Z., Lin, S., Hu, H.: Video swin transformer. In: Proceedings of the IEEE/CVF conference on computer vision and pattern recognition. pp. 3202--3211 (2022)

\bibitem{lyudvichenko2017semiautomatic}
Lyudvichenko, V., Erofeev, M., Gitman, Y., Vatolin, D.: A semiautomatic saliency model and its application to video compression. In: 2017 13th IEEE International Conference on Intelligent Computer Communication and Processing (ICCP). pp. 403--410. IEEE (2017)

\bibitem{lyudvichenko2019predicting}
Lyudvichenko, V., Vatolin, D.: Predicting video saliency using crowdsourced mouse-tracking data. In: Proceedings of the GraphiCon. pp. 127--130 (2019)

\bibitem{mahadevan2009spatiotemporal}
Mahadevan, V., Vasconcelos, N.: Spatiotemporal saliency in dynamic scenes. IEEE transactions on pattern analysis and machine intelligence  \textbf{32}(1),  171--177 (2009)

\bibitem{marat2009modelling}
Marat, S., Ho~Phuoc, T., Granjon, L., Guyader, N., Pellerin, D., Gu{\'e}rin-Dugu{\'e}, A.: Modelling spatio-temporal saliency to predict gaze direction for short videos. International journal of computer vision  \textbf{82}(3),  231--243 (2009)

\bibitem{mathe2014actions}
Mathe, S., Sminchisescu, C.: Actions in the eye: Dynamic gaze datasets and learnt saliency models for visual recognition. IEEE transactions on pattern analysis and machine intelligence  \textbf{37}(7),  1408--1424 (2014)

\bibitem{miangoleh2023realistic}
Miangoleh, S.M.H., Bylinskii, Z., Kee, E., Shechtman, E., Aksoy, Y.: Realistic saliency guided image enhancement. In: Proceedings of the IEEE/CVF Conference on Computer Vision and Pattern Recognition. pp. 186--194 (2023)

\bibitem{min2019tased}
Min, K., Corso, J.J.: Tased-net: Temporally-aggregating spatial encoder-decoder network for video saliency detection. In: Proceedings of the IEEE/CVF International Conference on Computer Vision. pp. 2394--2403 (2019)

\bibitem{mital2011clustering}
Mital, P.K., Smith, T.J., Hill, R.L., Henderson, J.M.: Clustering of gaze during dynamic scene viewing is predicted by motion. Cognitive computation  \textbf{3},  5--24 (2011)

\bibitem{aim2024vsrqa}
Molodetskikh, I., Borisov, A., Vatolin, D.S., Timofte, R., et~al.: {AIM} 2024 challenge on video super-resolution quality assessment: Methods and results. In: Proceedings of the European Conference on Computer Vision (ECCV) Workshops (2024)

\bibitem{salfom}
Moradi, M., Moradi, M., Rundo, F., Spampinato, C., Borji, A., Palazzo, S.: Salfom: Dynamic saliency prediction with video foundation models. arXiv preprint arXiv:2404.03097  (2024)

\bibitem{aim2024snr}
Nazarczuk, M., Catley-Chandar, S., Tanay, T., Shaw, R., Pérez-Pellitero, E., Timofte, R., et~al.: {AIM} 2024 sparse neural rendering challenge: Methods and results. In: Proceedings of the European Conference on Computer Vision (ECCV) Workshops (2024)

\bibitem{aim2024snr_dataset}
Nazarczuk, M., Tanay, T., Catley-Chandar, S., Shaw, R., Timofte, R., Pérez-Pellitero, E.: {AIM} 2024 sparse neural rendering challenge: Dataset and benchmark. In: Proceedings of the European Conference on Computer Vision (ECCV) Workshops (2024)

\bibitem{papoutsaki2017searchgazer}
Papoutsaki, A., Laskey, J., Huang, J.: Searchgazer: Webcam eye tracking for remote studies of web search. In: Proceedings of the 2017 conference on conference human information interaction and retrieval. pp. 17--26 (2017)

\bibitem{patel2021saliency}
Patel, Y., Appalaraju, S., Manmatha, R.: Saliency driven perceptual image compression. In: Proceedings of the IEEE/CVF winter conference on applications of computer vision. pp. 227--236 (2021)

\bibitem{Payne2023}
Payne, K.: Online Mouse Tracking as a Measure of Attention in Videos, Using a Mouse-Contingent Bi-Resolution Display. Thesis, Department of Psychological Sciences (December 2023), \url{https://hdl.handle.net/2097/43472}, master of Science

\bibitem{riche2013saliency}
Riche, N., Duvinage, M., Mancas, M., Gosselin, B., Dutoit, T.: Saliency and human fixations: State-of-the-art and study of comparison metrics. In: Proceedings of the IEEE international conference on computer vision. pp. 1153--1160 (2013)

\bibitem{rudoy2012crowdsourcing}
Rudoy, D., Goldman, D.B., Shechtman, E., Zelnik-Manor, L.: Crowdsourcing gaze data collection. arXiv preprint arXiv:1204.3367  (2012)

\bibitem{aim2024cvqa}
Smirnov, M., Gushchin, A., Antsiferova, A., Vatolin, D.S., Timofte, R., et~al.: {AIM} 2024 challenge on compressed video quality assessment: Methods and results. In: Proceedings of the European Conference on Computer Vision (ECCV) Workshops (2024)

\bibitem{tavakoli2017saliency}
Tavakoli, H.R., Ahmed, F., Borji, A., Laaksonen, J.: Saliency revisited: Analysis of mouse movements versus fixations. In: Proceedings of the ieee conference on computer vision and pattern recognition. pp. 1774--1782 (2017)

\bibitem{tavakoli2019dave}
Tavakoli, H.R., Borji, A., Rahtu, E., Kannala, J.: Dave: A deep audio-visual embedding for dynamic saliency prediction. arXiv preprint arXiv:1905.10693  (2019)

\bibitem{vig2014large}
Vig, E., Dorr, M., Cox, D.: Large-scale optimization of hierarchical features for saliency prediction in natural images. In: Proceedings of the IEEE conference on computer vision and pattern recognition. pp. 2798--2805 (2014)

\bibitem{wang2018revisiting}
Wang, W., Shen, J., Guo, F., Cheng, M.M., Borji, A.: Revisiting video saliency: A large-scale benchmark and a new model. In: Proceedings of the IEEE Conference on Computer Vision and Pattern Recognition (2018)

\bibitem{wang2019youtube}
Wang, Y., Inguva, S., Adsumilli, B.: Youtube ugc dataset for video compression research. In: 2019 IEEE 21st International Workshop on Multimedia Signal Processing (MMSP). pp.~1--5. IEEE (2019)

\bibitem{xiong2023casp}
Xiong, J., Wang, G., Zhang, P., Huang, W., Zha, Y., Zhai, G.: Casp-net: Rethinking video saliency prediction from an audio-visual consistency perceptual perspective. In: Proceedings of the IEEE/CVF Conference on Computer Vision and Pattern Recognition. pp. 6441--6450 (2023)

\bibitem{xu2015turkergaze}
Xu, P., Ehinger, K.A., Zhang, Y., Finkelstein, A., Kulkarni, S.R., Xiao, J.: Turkergaze: Crowdsourcing saliency with webcam based eye tracking. arXiv preprint arXiv:1504.06755  (2015)

\bibitem{yang2019sgdnet}
Yang, S., Jiang, Q., Lin, W., Wang, Y.: Sgdnet: An end-to-end saliency-guided deep neural network for no-reference image quality assessment. In: Proceedings of the 27th ACM international conference on multimedia. pp. 1383--1391 (2019)

\bibitem{zhang2021sa}
Zhang, Q.L., Yang, Y.B.: Sa-net: Shuffle attention for deep convolutional neural networks. In: ICASSP 2021-2021 IEEE International Conference on Acoustics, Speech and Signal Processing (ICASSP). pp. 2235--2239. IEEE (2021)

\bibitem{zhang2017study}
Zhang, W., Liu, H.: Study of saliency in objective video quality assessment. IEEE Transactions on Image Processing  \textbf{26}(3),  1275--1288 (2017)

\bibitem{zhou2023transformer}
Zhou, X., Wu, S., Shi, R., Zheng, B., Wang, S., Yin, H., Zhang, J., Yan, C.: Transformer-based multi-scale feature integration network for video saliency prediction. IEEE Transactions on Circuits and Systems for Video Technology  \textbf{33}(12),  7696--7707 (2023)

\end{thebibliography}
\end{document}